\documentclass[journal]{IEEEtran}

\usepackage{caption}
\usepackage{graphicx}
\usepackage{amsmath,amssymb} 
\usepackage{color}
\usepackage{multirow}
\usepackage{mathrsfs}
\usepackage{subfigure}
\usepackage{array}
\usepackage{threeparttable}
\usepackage{float}
\ifCLASSINFOpdf

\else

\fi

\begin{document}

\title{A Novel Transferability Attention Neural Network Model for EEG Emotion Recognition}

\author{Yang~Li, Boxun Fu,
	Fu Li$^*$, Guangming Shi, Wenming~Zheng

\thanks{Yang Li, Boxun Fu, Fu Li and Guangming Shi are with the Key Laboratory of Intelligent Perception and Image Understanding of Ministry of Education, the School of Artificial Intelligence, Xidian University, Xi’an, 710071, China.\it{($^*$Corresponding author: Fu Li (E-mail: fuli@mail.xidian.edu.cn).)} \protect}
\thanks{Wenming Zheng is with the Key Laboratory of Child Development and Learning Science (Ministry of Education), School of Biological Sciences and Medical Engineering,
Southeast University, Nanjing, Jiangsu, 210096, China.\protect }
}

\markboth{}
{Shell \MakeLowercase{\textit{et al.}}: Bare Demo of IEEEtran.cls for Journals}

\maketitle
\begin{abstract}
	The existed methods for electroencephalograph (EEG) emotion recognition always train the models based on all the EEG samples indistinguishably. However, some of the source (training) samples may lead to a negative influence because they are significant dissimilar with the target (test) samples. So it is necessary to give more attention to the EEG samples with strong transferability rather than forcefully training a classification model by all the samples. Furthermore, for an EEG sample, from the aspect of neuroscience, not all the brain regions of an EEG sample contains emotional information that can transferred to the test data effectively. Even some brain region data will make strong negative effect for learning the emotional classification model. Considering these two issues, in this paper, we propose a transferable attention neural network (TANN) for EEG emotion recognition, which learns the emotional discriminative information by highlighting the transferable EEG brain regions data and samples adaptively through local and global attention mechanism. This can be implemented by measuring the outputs of multiple brain-region-level discriminators and one single sample-level discriminator. We conduct the extensive experiments on three public EEG emotional datasets. The results validate that the proposed model achieves the state-of-the-art performance.
\end{abstract}


\IEEEpeerreviewmaketitle


\section{Introduction}
Emotion plays an important role in human daily life. It influences our rational decision-making, perception and cognition, and is essential in interpersonal communication~\cite{picard2000affective}. Thus, it is necessary to make machines to understand human emotions in the field of human-computer interaction (HCI). To this end, the technology of emotion recognition provides a possible way for computers to capture human emotions, which is the first step to improve and humanize the interaction between humans and machines.

Generally, emotion recognition measures the emotional states by analyzing the data of bodily reactions under emotional conditions~\cite{garcia2019review}. These reactions, including speech, facial expression and gesture, can adequately express our emotions under most circumstances. Nevertheless, these methods are subjective and cannot guarantee the authenticity of emotion~\cite{chen2019accurate}. Except for the above external methods, the internal physiological variables tend to be much close to the real emotions. Human brain, as the source of all the reactions, can reflect the mental activities including the emotion states. According to the studies of neurophysiology and psychology, EEG has the ability to record the brain neural activities, and can be used to decode the effective information of human emotional states~\cite{sammler2007music,mathersul2008investigating}. Consequently, EEG emotion recognition has received substantial attention from human-computer interaction and pattern recognition research communities in recent years~\cite{lin2010eeg,zheng2015investigating,li2019eeg}.

Most EEG emotion recognition methods focus on two major tasks, i.e., EEG feature extraction and classification. The first task aims at seeking the discriminative emotion-related information from the raw EEG signals. EEG emotional signals usually consist of many neural processes and hence present a highly heterogeneous and nonstationary behavior~\cite{garcia2019review}. Hence, how to extract the specific emotion information that contribute to the emotion recognition becomes a very important task. In~\cite{jenke2014feature}, Jenke et al. summarized and evaluated all the existing EEG features extracted from time domain, frequency domain and time-frequency domain on their self-recorded EEG emotional dataset. The target of classification is modeling the correlation between the EEG emotional feature and the class labels, which leads to the interpretation of raw EEG emotional signals. Classification performance provides insight about how well a trained model can estimate the emotional state. Many advanced classification algorithms have been proposed over the years. For example, Zheng et al.~\cite{zheng2017multichannel} proposed a group sparse canonical correlation analysis method for simultaneous EEG channel selection and emotion recognition. Li et al.~\cite{li2019eeg} fused the information propagation patterns and activation difference in the brain to improve emotional recognition. In~\cite{alarcao2017emotions}, Alarcao and Fonseca summarized, reviewed and compared these works comprehensively. 

%
Recently, many domain adaptation methods have been proposed to deal with EEG emotion recognition, especially in the subject-independent task, where the source and target data come from different subjects. These methods have significantly advanced the EEG emotion recognition task. For example, Zheng and Lu~\cite{zheng2016personalizing} evaluated four different domain adaptation approaches including Transfer component analysis (TCA)~\cite{pan2011domain}, Kernel Principle Analysis (KPCA)~\cite{scholkopf1998nonlinear}, Transductive Support Vector Machine (T-SVM)~\cite{collobert2006large} and Transductive Parameter Transfer (TPT)~\cite{sangineto2014we} on SEED dataset, and find that the accuracy can be improved by 20$\%$ compared with the generic classifier. Lan et al.~\cite{lan2018domain} made a comparative study on several state-of-the-art domain adaptation techniques on two EEG emotional datasets and the experiment results show that using domain adaptation technique can improve the accuracy significantly by 7.25$\%$ and 13.40$\%$ compared with the baseline accuracy where no domain adaptation technique is used. In all the domain adaptation methods, the most well-established one is the domain adversarial neural network (DANN)~\cite{ganin2016domain}, which constructs a two players mini-max game by using a domain discriminator that works adversarially with the feature extractor to generate the domain-invariable data representations. Li et al. adopted this setting and proposed a bi-hemisphere domain adversarial neural network (BiDANN) for EEG emotion recognition and achieved the state-of-the-art performance~\cite{li2018bi}.

Nevertheless, we argue that the there are two issues need to be better addressed in EEG emotion recognition tasks. The first one is how to identify the positive EEG samples that consist of more emotion-related information. EEG emotional signals usually consist of many neural processes and are much vulnerable to negative effect of irrelevant knowledge, which incurs that some training EEG samples are significantly dissimilar with the test ones. Exploring how to highlight the positive EEG emotional samples and weaken the effect of negative samples will contribute more to emotion recognition. The second issue is how to weight the variability of different brain regions for EEG emotion recognition. Some studies of neuroscience have shown that different brain regions have different contributions for emotion expression~\cite{kragel2016decoding}. In an EEG emotional sample, it is obvious that not all the brain regions contain the knowledge of emotion that can be transferred to the test samples. Making a strategy to distinguish the transferable and nontransferable brain regions is helpful to improve EEG emotion recognition.

To this end, in this paper, we propose a transferable attention neural network (TANN) to deal with the above tranferability learning problem for EEG emotion recognition. This transferability of data can be measured by calculating from the outputs of domain discriminators. Specifically, for the domain adversarial neural network~\cite{ganin2016domain}, the output of domain discriminator is the probability of input data belongs to source or target domain. When the probability approaches 0, it represents the input data belongs to source domain, while approaching 1 indicates that it belongs to the target domain. Therefore, TANN takes advantages of the domain discriminator to measure the transferability from the training data to test data. Concretely, the framework of TANN includes the following three major modules:
\begin{itemize}
	\item \textbf{Feature extractor.} The goal of feature extractor is to extract the high-level discriminative deep feature from raw EEG data for classification. EEG data is made of several electrodes that are set under the coordinates on the scalp, which are predefined referring to the locations of different brain regions. In the feature learning procedure, we should well retain this intrinsic structural information that will be helpful for classification. To achieve this, TANN employs two directional recurrent neural networks (RNN) that traverse all the electrodes from horizontal and vertical directions, which will construct a complete relationship and generate discriminative deep features for all the EEG electrodes. 	
	\item \textbf{Attention module.} The attention module aims to weight the input training data according to the level of transferability. For EEG emotional data, there is a large distribution gap between training and test data, resulting that some training EEG data are significantly dissimilar with the test. Moreover, from the aspect of neuroscience, not all the brain regions of an EEG sample contains emotional information that can transferred to the test data effectively. Therefore, TANN employs multiple brain-region-level and one sample-level discriminators to assess the transferability of EEG sample and the inside brain region data, then strengthen or weaken the contributions of these brain regions and samples for emotion classification.
	
%
%
%
%
%
%
%
%
	\item \textbf{Classifier.} Like most supervised learning methods, we introduce a classifier to predict the emotion class label based on the deep features obtained by the feature extractor. It will guide the feature extracting process towards generate more discriminative EEG features for emotion classification.
\end{itemize}

To the best of our knowledge, this is the first work to exploit the global and local transferability of EEG signals for emotion recognition. The experimental results verify the proposed TANN method can achieve the state-of-the-art performance on three public datasets.


%
%
%
%
%
%

\section{Preliminary}

In this section, we briefly overview the preliminary of transferable attention and then address how we can apply it to EEG emotion recognition.

Most attention based methods focus on how to highlight or weaken different parts in an image according to their contribution for classification but neglect the evaluation for each training sample~\cite{vaswani2017attention}. It is known that not all the training samples are similar with the test. It will be a negative influence in the learning process if we feed the model with all the training samples forcefully. Transferable attention (TA) is designed to deal with this problem~\cite{wang2019transferable}. When a training sample is much easier to be transferred to the test, it will be rewarded with more attention due to the high similarity with the test data, which is called transferable attention. Inspired by adversarial learning methods, this attention can be realized by calculating the outputs of the discriminator, which can reflect the similarity between training and test data. 

Since in EEG emotion recognition tasks, not all the training EEG data are useful in the process of learning a model, exploring the transferability of EEG data will be meaningful and can further improve EEG emotion recognition.
%
%

\section{The proposed model for EEG emotion recognition}
\begin{figure*}[htb]
	\centering
	\includegraphics[width=1.75\columnwidth]{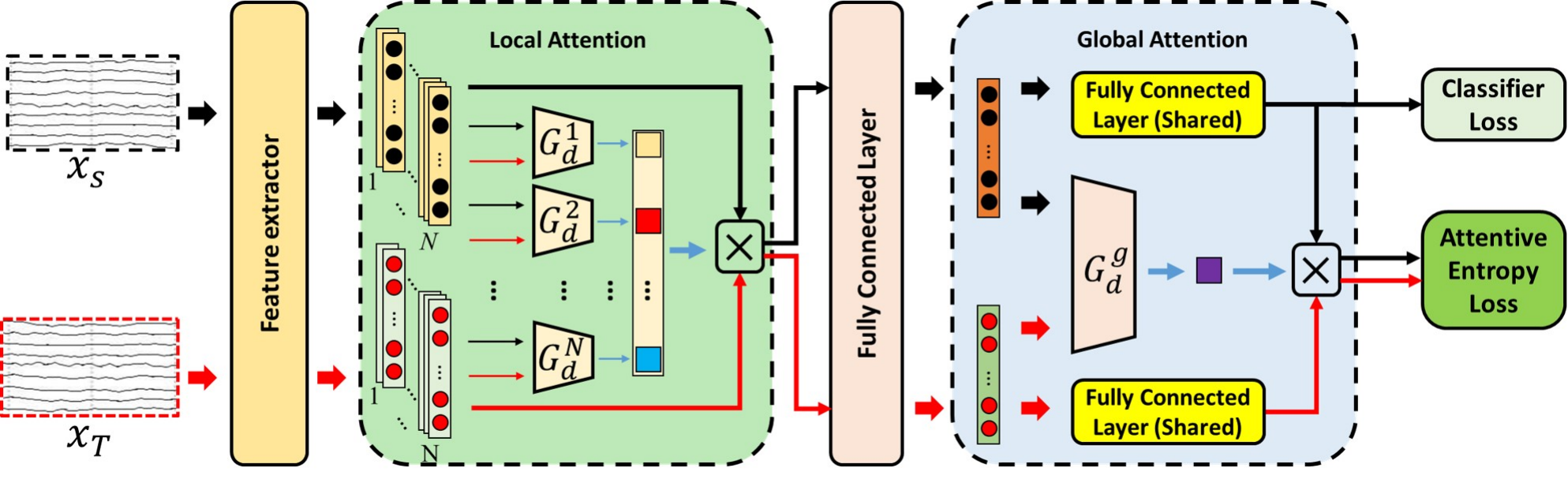} \\
	\caption{The framework of TANN. TANN consists of two major modules, i.e., local and global attentions, that can make the model focus on the brain regions and samples with higher transferability.}
	\label{Fig: TANN framework}
\end{figure*}
To specify the proposed method clearly, we illustrate the framework of the proposed TANN model in Fig.~\ref{Fig: TANN framework}. TANN aims to distinguish which training samples are easy or hard to be transferred to test samples. Through penalizing these training samples, it can further improve EEG emotion recognition. Besides, considering not all the brain regions have the equal transferability, as well as measuring the similarity across EEG samples, TANN also focuses on the brain regions with high transferability. To achieve this goal, we adopt local and global attentions to the EEG emotion sample and its inside brain regions' data, respectively. These attention weights can be obtained from the outputs of multiple local and one global domain discriminators. Concretely, TANN consists of three major modules, i.e., feature extractor, attention layers, and classifier. In the following, we illustrate these parts detailedly.

 
\label{Sec: The proposed method}

\subsection{Feature extractor}
\begin{figure}[htb]
	\centering
	\includegraphics[width=1\columnwidth]{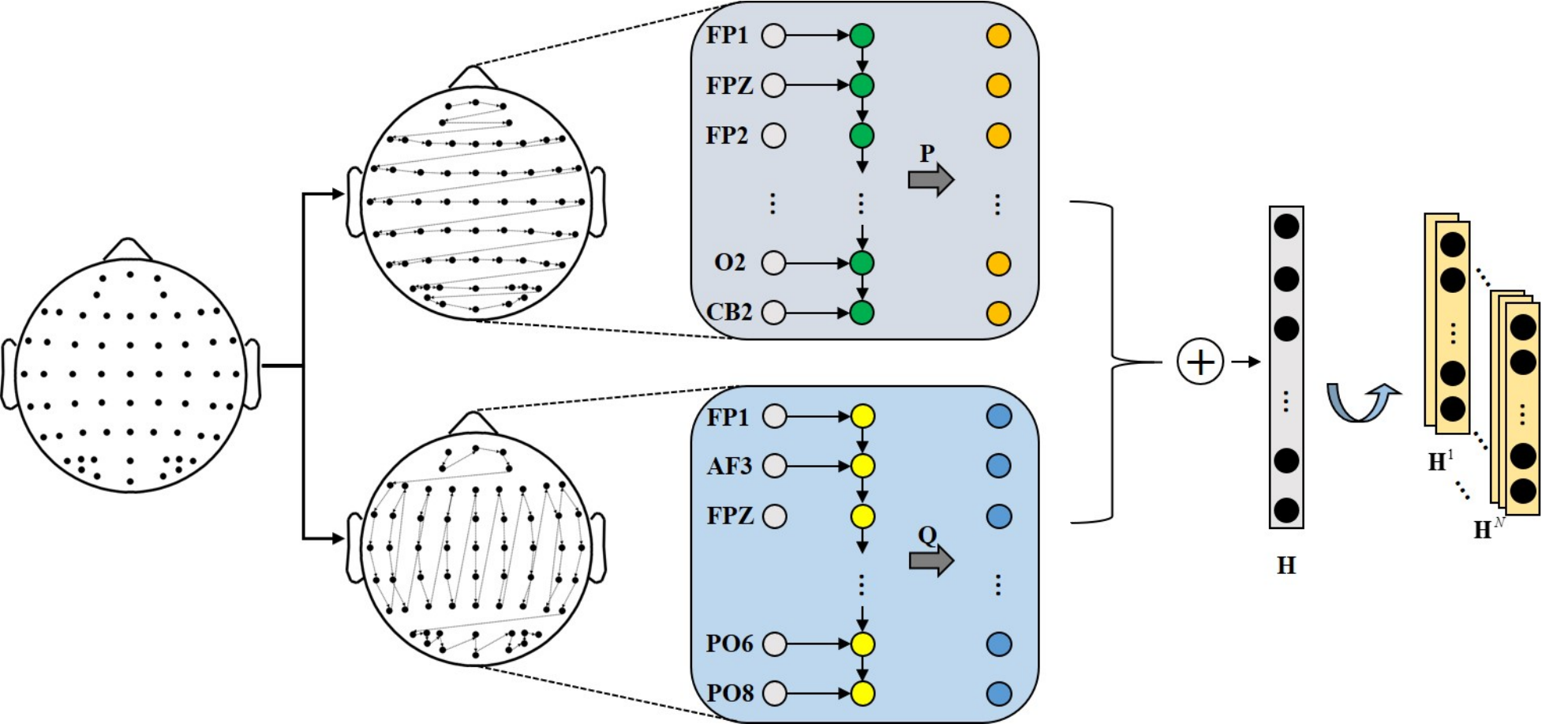} \\
	\caption{The process of feature extraction. We first extract the deep feature for each electrode, and then rearrange them to form the data representation of brain regions.}
	\label{Fig: horizontal and vertival}
\end{figure}

The process of feature extraction is depicted in Fig.~\ref{Fig: horizontal and vertival}, and the goal is to represent the EEG emotional data in a more discriminative feature space so as to improve the EEG classification performance. The EEG deep features are extracted by two directional RNN modules that traverse the spatial regions under two predefined stacks, which are determined with respect to horizontal and vertical directions. These two directional RNNs are complementary to construct a complete relationship of electrodes’ locations that avoid losing the intrinsic structural information of EEG data. By doing this, we can obtain the high-level features for each EEG electrode that facilitate to construct the brain regions' features.


Concretely, for an EEG sample $\mathbf{X}=[\mathbf{x}_1,\cdots,\mathbf{x}_n]\in \mathbb{R}^{d\times n}$, where $d$ and $n$ are the dimension and number of EEG electrode, the above process can be formulated as
\begin{align}
\begin{aligned}
\mathbf{s}^h_i = \sigma(\mathbf{U}^h\mathbf{x}^h_i + \sum\nolimits_{j=1}^n e^h_{ij}\mathbf{V}^h \mathbf{h}^h_j + \mathbf{b}^h) \in \mathbb{R}^{d_f},\\
e^h_{ij} = \left\{
\begin{array}{lr}
1, & \textrm{if $\mathbf{x}^h_j$ $\in$ $\mathcal{N}$ ($\mathbf{x}^h_i$) },\\
0, & \textrm{otherwise},~~~~~~~
\end{array}~~~~~~~~
\right.
\end{aligned}
\end{align}	
\begin{align}
\begin{aligned}
\mathbf{s}^v_i = \sigma(\mathbf{U}^v\mathbf{x}^v_i + \sum\nolimits_{j=1}^n e^v_{ij}\mathbf{V}^v \mathbf{h}^v_j + \mathbf{b}^v) \in \mathbb{R}^{d_f},\\
e^v_{ij} = \left\{
\begin{array}{lr}
1, & \textrm{if $\mathbf{x}^v_j$ $\in$ $\mathcal{N}$ ($\mathbf{x}^v_i$) },\\
0, & \textrm{otherwise},~~~~~~~~
\end{array}~~~~~~~~
\right.
\end{aligned}
\end{align}	
where $\mathbf{s}^{\cdot}_i$ is the hidden unit of the RNN module as well as the data representation for the electrode $\mathbf{x}_i$, and $d_f$ is its dimension; $\{\mathbf{U}^{\cdot} \!\in\! \mathbb{R}^{d_f \times d}$, $\mathbf{V}^{\cdot} \!\in\! \mathbb{R}^{d_f \times d_f}$, $\mathbf{b}^{\cdot} \!\in\! \mathbb{R}^{d_f \times 1}\}$ are the learnable transformation matrices of RNN module; $\sigma(\cdot)$ denotes the nonlinear operation such as Sigmoid function; and $\mathcal{N}(\mathbf{x}^{\cdot}_i$) denotes the set of predecessors of node $\mathbf{x}^{\cdot}_i$. 

Due to that TANN consists of horizontal and vertical directional RNNs to represent EEG electrode, we can obtain the data representations that not only contain the information of the electrodes itself but also the nearby relationship. Specifically, it can be expressed as $\mathbf{S}^h = \{\mathbf{s}^h_i\}$ that contains the information from left and right electrodes, and $\mathbf{S}^v = \{\mathbf{s}^v_i\}$ that includes the information from up and down electrodes. To integrate these spatial information into a overall representation, we arrange the order of the columns of $\mathbf{S}^h$ and $\mathbf{S}^v$, and use two transformation matrices $\mathbf{P}$ and $\mathbf{Q}$ to obtain the deep features $\mathbf{H}=\{\mathbf{h}_k\}$ for all the electrodes, in which
 \begin{eqnarray}
 \mathbf{h}_i = \mathbf{P}\mathbf{s}^h_i + \mathbf{Q}\mathbf{s}^v_i + \mathbf{b}\in \mathbb{R}^{ d_{f'}},  i \in \{1,\cdots,n\}.
 \end{eqnarray}
Here $\mathbf{h}_i$ is the deep representation of electrode $\mathbf{x}_i$ that kepdf the location structural relation, $d_f$ is the dimension.

\subsection{Attention layers}
%

For EEG emotion samples, there is a large distribution gap between training and test data. Some training samples are very dissimilar with the test ones. Therefore, to avoid training a model with all the source samples indiscriminately, TANN measures the transferability of all the training samples and then strengthen or weaken them in the learning process of the model. Besides, as we know, for emotion recognition, 
not all the brain regions of an EEG sample contains emotional information that can transferred to the test data effectively. Some brain regions are more transferable than the others. Due to this, TANN not only employs a global attention layer to weight the sample-level transferability but also a local attention layer as a complement to focus on the brain-region-level transferability. Specifically, the transferability is quantified by the entropy of the outputs of domain discriminator. The domain discriminator can generate the probability of confusion between source (training) and target (test) data. When the probability approaches 0.5, it indicates that the input has good ability to confuse the domain discriminator, which nicely meet our need to highlight the data with positive transferability. In the following, we will demonstrate how to achieve the local and global attentions by transferability learning.

\subsubsection{Local transferable attention on brain-region-level}
After obtaining the data representation $\mathbf{h}_i$ of each electrode of $\mathbf{X}$, TANN employs local attention to highlight the brain regions with high transferability. Here we first group the electrodes into several clusters according to the associated brain region locations, which can be formulated as 
\begin{align}
\begin{aligned}
\label{Eq: Brain Region}
\mbox{brain region 1:}~& \mathbf{H}^1=[\mathbf{h}^1_{1},\mathbf{h}^1_{2},\cdots,\mathbf{h}^1_{n_1}],\\
\cdots ~~~~~~~& ~~~~~~~~\cdots \\
\mbox{brain region N:}~&  \mathbf{H}^N=[\mathbf{h}^N_{1},\mathbf{h}^N_{2},\cdots,\mathbf{h}^N_{ n_N}],
\end{aligned}
\end{align}
where $N$ is the number of brain regions, $n_c$ denotes the number of electrodes in the $c$-th brain region, $n_1+\cdots+n_N=n$. In this case, the reordered deep feature can be expressed as
\begin{eqnarray}\label{Eq: Reordered Deep Feature}
\hat{\mathbf{H}} = [\mathbf{H}^1,\cdots,\mathbf{H}^N].
\end{eqnarray}

Based on the above process, we can obtain the deep features of all the brain regions from source and target EEG samples, which can be denoted as $\hat{\mathbf{H}}_S= [\mathbf{H}^1_S,\cdots,\mathbf{H}^N_S]$ and $\hat{\mathbf{H}}_T= [\mathbf{H}^1_T,\cdots,\mathbf{H}^N_T]$. Then they are fed to $N$ local discriminators to calculate the transferability. Concretely, let $\mathbf{d}^{N_i}=\{d^{N_i}_s,d^{N_i}_t\}$ denote the output probability of one discriminator for brain region $N_i$, where $d^{N_i}_s$ and $d^{N_i}_t$ are the probabilities that the input belongs to the source and target data, respectively. Then we can quantity the transferability of this brain region through the entropy function in information theory~\cite{wang2019transferable}, which is defined as
\begin{eqnarray}\label{Eq: Entropy}
H(\mathbf{d}^{N_i}) = - d^{N_i}_s \cdot \log(d^{N_i}_s) - d^{N_i}_t \cdot \log(d^{N_i}_t).
\end{eqnarray}
Then the higher transferability of a brain region has, the more attention value is.

However, for an EEG signal, the emotion information is the most difficult component to be transferred. Due to this, we reverse the attention values for the brain regions to make the model pay attention on the difficult transferred brain regions. Thus the attention value for brain region $N_i$ is defined as
\begin{eqnarray}\label{Eq: 1-Entropy}
w^{N_i} = 1 - H(\mathbf{d}^{N_i}).
\end{eqnarray}

Besides, to mitigate the negative effect of wrong attentions, we adopt the residual attention mechanism to make the model more robust. Thus, after local attention layer, the data representations for EEG sample $\mathbf{X}$ can be formulated as  
\begin{eqnarray}\label{Eq: Local Attention Feature}
\hat{\mathbf{H}}' = [(1+w^1)\mathbf{H}^1,\cdots,(1+w^N)\mathbf{H}^N] \in \mathbb{R}^{ d_{f'}\times n }.
\end{eqnarray}

Here the loss function of the local discriminators for all the brain regions can be formulated as
\begin{eqnarray}
L^l_d = \frac{1}{N} \sum^N_{N_i=1} L^{l_{N_i}}_d(\mathbf{X}^S,\mathbf{X}^T |\theta^{l_{N_i}}_d),
\end{eqnarray}
where 
\begin{eqnarray}
L^{l_{N_i}}_d 
= -\sum\nolimits_{t=1}^{M_1} \textrm{log} p(0|\mathbf{X}^{S_{N_i}}_t) - \sum\nolimits_{t'=1}^{M_2} \textrm{log} p(1|\mathbf{X}^{T_{N_i}}_{t'})
\end{eqnarray}
denote the loss of the local discriminator for brain region $N_i$; $p(0|\mathbf{X}^{S_{N_i}}_t)$ and $p(1|\mathbf{X}^{T_{N_i}}_{t'})$ are the probabilities of the input data belongs to source and target domains respectively; $\theta^{l_{N_i}}_d$ is the parameter of the local attention network; 
$\mathbf{X}^{S_{N_i}}_t$ and $\mathbf{X}^{T_{N_i}}_{t'}$ represent the $N_i$ brain region data of the $t$-th and $t'$-th source and target sample, respectively; $M_1$ and $M_2$ are the number of the source and target data.

\subsubsection{Global transferable attention on sample-level}
Although the above local attention for all the brain regions can make a fine-grained transfer learning between the source and target domain data, there is a possible that the local domain discriminator find fewer brain regions to transfer. Meanwhile, due to the distribution difference, there are some negative samples in the source data that are very dissimilar with the target data. It will weak the efficiency If we force training the model with these negative samples equaling with the other positive samples. Hence, after weighting the transferability of brain regions with local attention, we adopt the global transferable attention on the sample-level to transfer the knowledge from source to target domain. 

Concretely, after local attention module, the input feature can be expressed as
\begin{eqnarray}\label{Eq: Global Attention Feature}
\tilde{\mathbf{H}} =  \hat{\mathbf{H}}' \mathbf{S} \in \mathbb{R}^{ d_{f'}\times n' },
\end{eqnarray}
where $\mathbf{S}$ is a learnable transformation matrix. Then it is sent to a global discriminator 
\begin{eqnarray}
L^g_d(\mathbf{X}^S,\mathbf{X}^T |\theta^g_d) = -\sum\nolimits_{t=1}^{M_1} \textrm{log} p(0|\mathbf{X}^S_t) ~~~~\nonumber\\
- \sum\nolimits_{t'=1}^{M_2} \textrm{log} p(1|\mathbf{X}^T_{t'}),
\end{eqnarray}
to highlight the EEG samples with higher transferability, where $\theta^g_d$ is the parameter of the global attention network. Concretely, let $\mathbf{d}=\{d_s,d_t\}$ denote the output probability of the global discriminator, where $d_s$ and $d_t$ are the probabilities that the input belongs to the source and target data respectively. The global attention value $w$ can be calculated as
\begin{eqnarray}\label{Eq: Global Attention}
w=1+H(\mathbf{d}), ~~~~~~~~~~~~~~~~~~~~~~~~\\
H(\mathbf{d}) = -d_s \cdot \log(d_s) - d_t \cdot \log(d_t).
\end{eqnarray}
Here we also adopt the residual mechanism to avoid the wrong attention. In this case, we obtain that the more transferability is, the larger attention value $w$ is.

Inspired by Long et al.~\cite{long2016unsupervised}, the entropy minimization principle can refine the classifier adaptation, which can increase the confidence of the classifier prediction. Thus, we utilize the global domain discriminator to generate the global attention values acting on the label entropy to enhance the certainty of the source samples that are more similar with the target samples. Then $w$ is embedded into the label entropy loss to achieve the function for global attention. Hence the loss function of the label entropy, which is called attentive entropy loss, can be written as
\begin{eqnarray}
\label{Eq: label entropy loss}
L_e(\mathbf{X}^S, \mathbf{X}^T | \theta_e) = \sum_{k=1}^{M_1+M_2} \sum_{c=1}^C-w \cdot p(c|\mathbf{X}_k) \cdot \textrm{log} p(c|\mathbf{X}_k),
\end{eqnarray}
where $\mathbf{X}_k$ is the $k$-th sample in $\{\mathbf{X}^S, \mathbf{X}^T\}$; $w$ is the global attention value for EEG sample $\mathbf{X}_k$; and $C$ is the number of emotion classes.

\subsection{Classifier}
To enhance the discriminative ability of the model, we add the classifier to TANN model. Concretely, based on the final feature vector $\tilde{\mathbf{H}}$ in Eq.~(\ref{Eq: Global Attention Feature}), we first arrange the matrix $\tilde{\mathbf{H}}$ into a vector $\tilde{\mathbf{h}}$, and then use the simple linear transform approach to predict the class label, which can be formulated as
\begin{eqnarray}\label{Eq: Classifier}
\mathbf{O} = \mathbf{G}\tilde{\mathbf{h}} +\mathbf{b}_c = [o_1,\cdots,o_C],
\end{eqnarray}
where $\mathbf{G} $ and $\mathbf{b}^c$ are the transformation matrices. Finally, the output vector $\mathbf{O}$ is fed into the softmax layer for emotion classification, which can be written as
\begin{eqnarray}
\label{Eq: classifier}
p(c|\mathbf{X}_t)= \exp(o_c)/\sum\nolimits^{C}_{i=1} \exp(o_i),
\end{eqnarray}  
where $p(c|\mathbf{X}_t)$ denotes the predicted probability that the input sample $\mathbf{X}_t$ belongs to the $c$-th class. As a result, the label $\tilde{l}$ of sample $\mathbf{X}_t$ is predicted as 
\begin{eqnarray}
\tilde{l} = arg \max_c  p(c|\mathbf{X}_t).
\end{eqnarray}  

Hence, the loss function of the classifier can be expressed as
\begin{align}
\begin{aligned}
\label{Eq: classifier loss}
L_c(\mathbf{X}^S |\theta_c) = \sum_{t=1}^{M_1} \sum_{c=1}^C-\tau (l, c)\cdot \textrm{log} p(c|\mathbf{X}_t),\\
\tau(l, c)=\left\{
\begin{array}{lr}
1, & \textrm{if $l$ =}~c,~~\\
0, & \textrm{otherwise},
\end{array}
\right.
\end{aligned}
\end{align}
where $\theta_c$ denotes the parameter of the classifier.

\subsection{The optimization}
In summary, the overall loss function includes four parts, i.e., local and global discriminator losses, classifier loss and the attentive entropy loss. Concretely, the loss function of the proposed TANN method can be formulated as
\begin{eqnarray}
\label{Eq: Overall loss function}
L(\mathbf{X}^S,\mathbf{X}^T |\theta_c,\theta_e,\theta^l_d,\theta^g_d) = L_c(\mathbf{X}^S |\theta_c) + \alpha L_e(\mathbf{X}^S,\mathbf{X}^T |\theta_e) \nonumber \\  - \beta ( \frac{1}{N} \sum^N_{N_i=1} L^{l_{N_i}}_d(\mathbf{X}^S,\mathbf{X}^T |\theta^{l_{N_i}}_d) + L^g_d(\mathbf{X}^S,\mathbf{X}^T |\theta^g_d)),
\end{eqnarray}
where $\alpha$ and $\beta$ are the hyper-parameters, $L^{l_{N_i}}_d$ and $L^g_d$ represent the losses of local and global attention discriminators. Then we iteratively optimize the classifier, attentive entropy, local and global attention discriminators. Concretely, the parameters can be found through minimizing and maximizing
\begin{eqnarray}
\label{Eq: Each loss function}
(\hat{\theta}_f,\hat{\theta}_c) \!\!\!\!&=&\!\!\!\! \arg \min_{\theta_f,\theta_c} L_c(\mathbf{X}^S |\theta_f,\theta_c,\hat{\theta}_e,\hat{\theta}^l_d,\hat{\theta}^g_d),\\
\hat{\theta}_e &=&\!\!\!\! \arg \min_{\theta_e} L_e(\mathbf{X}^S,\mathbf{X}^T |\hat{\theta}_f,\hat{\theta}_c,\theta_e,\hat{\theta}^l_d,\hat{\theta}^g_d),\\ 
\label{Eq: local discriminator loss function}
\hat{\theta}^{l_{N_i}}_d \!\!\!\!&=&\!\!\!\! \arg \max_{\theta^{l_{N_i}}_d} L^{l_{N_i}}_d(\mathbf{X}^S,\mathbf{X}^T  |\hat{\theta}_f,\hat{\theta}_c,\hat{\theta}_e,\theta^{l_{N_i}}_d,\hat{\theta}^g_d),~\\
\label{Eq: global discriminator loss function}
\hat{\theta}^g_d &=&\!\!\!\! \arg \max_{\theta^g_d} L^g_d(\mathbf{X}^S,\mathbf{X}^T |\hat{\theta}_f,\hat{\theta}_c,\hat{\theta}_e,\hat{\theta}^l_d,\theta^g_d).
\end{eqnarray}

The above maximization problem, i.e., Eq.~(\ref{Eq: local discriminator loss function}) and (\ref{Eq: global discriminator loss function}), can be transferred to a minimization problem through adopting a gradient reversal layer (GRL)~\cite{ganin2016domain} before the discriminator, which will act as an identity transform in the forward-propagation but reverse the gradient sign while performing the back-propagation operation. Then we can use the stochastic gradient decent (SGD) algorithm to solve the parameter optimization process easily. Specifically, the parameters can be updated by the rules below
\begin{eqnarray}
\theta_c \!\!&\leftarrow&\!\!\!\! \theta_c - \frac{\partial{L_c}}{\partial{\theta_c}},~~
\theta_e \leftarrow \theta_e - \alpha \cdot \frac{\partial{L_e}}{\partial{\theta_e}},~~\\
\theta^{l_{N_i}}_d \!\!\!\!\!\!&\leftarrow&\!\!\!\! \theta^{l_{N_i}}_d - \beta \cdot \frac{\partial{L^{l_{N_i}}_d}}{\partial{\theta^{l_{N_i}}_d}},~~
\theta^g_d \leftarrow \theta^g_d - \beta \cdot \frac{\partial{L^g_d}}{\partial{\theta^g_d}}, ~ \\
\theta_f \!\!&\leftarrow&\!\!\!\! \theta_f - (\frac{\partial{L_c}}{\partial{\theta_f}} + \alpha \cdot \frac{\partial{L_e}}{\partial{\theta_f}} - \beta \cdot \frac{\partial{L^{l_{N_i}}_d}}{\partial{\theta_f}}- \beta \cdot \frac{\partial{L^g_d}}{\partial{\theta_f}}).~~~
\end{eqnarray}

\section{Experiments}
\label{Sec: Experiment}
\subsection{Datasets and settings}


To evaluate the proposed TANN method adequately, we conduct the experiments on three public EEG emotion datasets, namely,
\begin{enumerate}
	\item[(1)] \textbf{SEED}~\cite{zheng2015investigating} dataset is a standard benchmark for EEG emotion recognition. It contains three types of emotions, i.e., \textit{happy}, \textit{neutral} and \textit{sad}, from 15 subjects' EEG emotional signals.
	
	\item[(2)] \textbf{SEED-IV}\footnote{Note that both SEED-IV and MPED are multi-modal datasets. MPED consists of 30 subjects' EEG data, among which 23 subjects contain multi-modal data. In this experiment, we only use the EEG modal data.\label{Footnote: SEED-IV}}~\cite{zheng2018emotionmeter} dataset includes four types of emotions from 15 subjects. Compared with SEED, it contains an extra emotion \textit{fear}.
	
	\item[(3)] \textbf{MPED}\textsuperscript{\ref {Footnote: SEED-IV}}~\cite{song2018eeg} dataset includes seven refined emotion types, i.e., \textit{joy}, \textit{funny}, \textit{neutral}, \textit{sad}, \textit{fear}, \textit{disgust} and \textit{anger} from 30 subjects.
\end{enumerate}

On these datasets, we design two kinds of EEG emotion recognition experiments including the subject-dependent and subject-independent ones. Table~\ref{Tabel: database} summarizes the number of training and test samples, and the experimental protocols used in the experiments. The concrete protocols are described as follows:
\begin{itemize}
	\item \textbf{The subject-dependent experiment} - In this experiment, the training and test data come from the same subject but different trials. We adopt the same protocols as~\cite{zheng2015investigating}, \cite{zheng2018emotionmeter} and \cite{song2019mped}. Namely, for SEED, we use the former nine trials of EEG data per session of each subject as source (training) domain data while using the remaining six trials per session as target (test) domain data; for SEED-IV, we use the first sixteen trials per session of each subject as the training data, and the last eight trials containing all emotions (each emotion with two trials) as the test data; for MPED, we use twenty-one trials of EEG data as training data and the rest seven trials consisting of seven emotions as test data for each subject. The mean accuracy (ACC) and standard deviation (STD) are used as the evaluation criteria for all the subjects in the dataset. 
	
	\item \textbf{The subject-independent experiment} - In this experiment, the training and test data come from different subjects, which is a harder task than the above subject-dependent one but more conductive to practical applications. We adopt the leave-one-subject-out (LOSO) cross-validation strategy~\cite{zheng2016personalizing} to evaluate the proposed TANN model. LOSO strategy uses the EEG signals of one subject as test data and the rest subjects' EEG signals as training data. This procedure is repeated such that the EEG signals of each subject will be used as test data once. Again, the mean accuracy (ACC) and standard deviation (STD) are used as the evaluation criteria.

\end{itemize}

Besides, we use the released handcraft features, namely, the differential entropy (DE) in SEED and SEED-IV, and the Short-Time Fourier Transform (STFT) in MPED, as the input to feed our model. Thus the sizes $d\!\times\! n$ of the input sample $\mathbf{X}_t$ are $5\!\times\!62$, $5\!\times\!62$ and $1\!\times\!62$ for these three datasets, respectively. Moreover, in the experiment, we respectively set the dimension $d_f$ and $d_f'$ of the feature extractor to 32; the number of brain region $N$ to 16\footnote{Concretely, the brain regions include Pre-Frontal (AF3, FP1, FPZ, FP2, AF4), Frontal (F3, F1,	FZ, F2, F4), Left Frontal (F7, F5), Right Frontal (F8, F6), Left Temporal (FT7, FC5, T7, C5, TP7, CP5), Right Temporal (FT8, FC6, T8, C6, TP8, CP6), Frontal Central (FC3, FC1, FCZ, FC2, FC4), Central (C3, C1, CZ, C2, C4), Central Parietal (CP3, CP1, CPZ, CP2, CP4), Left Parietal (P7, P5), Right Parietal (P8, P6), Parietal (P3, P1, PZ, P2, P4), Left Parietal Occipital (PO7, PO5, CB1), Right Parietal Occipital (PO8, PO6, CB2), Parietal Occipital (PO3, POZ, PO4), Occipital (O1, OZ, O2) lobes.}; the dimension $n'$ of the input for the global attention layer to 6; the hyper-parameters $\alpha$ and $\beta$ are both set to 0.1 throughout the experiment. Specifically, we implemented TANN using TensorFlow\footnote{https://www.tensorflow.org/} on one Nvidia 1080Ti GPU. The learning rate, momentum and weight decay rate are set as 0.003, 0.9 and 0.95, respectively. The network is trained using SGD with batch size of 200. 
\begin{table}
	\caption{The number of training and test samples, and the experimental protocols used in the experiment.}
	\label{Tabel: database}
	\centering
	\renewcommand{\arraystretch}{1.3}
	\subtable[The subject-dependent experiment]{
		\begin{threeparttable}	
			\begin{tabular}{|c|c|c|c|c|}
				\hline
				\multicolumn{2}{|c|}{\textbf{Dataset}}  &\textbf{Training} & \textbf{Test} &\textbf{Protocol} \\ 
				\hline
				\multicolumn{2}{|c|}{{SEED}}  &   2010   &  1384   & [Zheng and Lu]\cite{zheng2015investigating} \\
				\hline
				& Session 1  &   561    &  290    &\multirow{3}{*}{[Zheng et al.]\cite{zheng2018emotionmeter}}    \\
				SEED-IV          & Session 2  &  550    &  282    &   \\
				& Session 3 &   576    &  246    &   \\
				\hline
				\multicolumn{2}{|c|}{{MPED}}  &  2520   &  840    & [Song et al.]\cite{8606087} \\
				\hline
			\end{tabular}	
	\end{threeparttable}}
	\subtable[The subject-independent experiment]{
		\begin{threeparttable}	
			\begin{tabular}{|c|c|c|c|c|}
				\hline
				\multicolumn{2}{|c|}{\textbf{Dataset}}  &\textbf{Training} & \textbf{Test} &\textbf{Protocol} \\ 
				\hline
				\multicolumn{2}{|c|}{{SEED}}  &   47516   &  3394   & [Zheng et al.]\cite{zheng2016personalizing} \\
				\hline
				& Session 1  &   11914    &  851    &\multirow{3}{*}{LOSO$^*$}    \\
				SEED-IV & Session 2  &  11648    &  832    &   \\
				& Session 3 &   11508    &  822    &   \\
				\hline
				\multicolumn{2}{|c|}{{MPED}}  &  97440   &  3360    & LOSO$^*$ \\
				\hline
			\end{tabular}
	\begin{tablenotes}[para]
		\footnotesize $*$ LOSO denotes the leave-one-subject-out strategy.
	\end{tablenotes}	
	\end{threeparttable}}
\end{table}

\subsection{Experiment results}
To validate the classification superiority of TANN, we also conduct the same experiments using various existed methods. Recall that the distribution gap in the subject-independent task is much larger than that in the subject-dependent one. In this case, domain adaptation methods shall be properly employed in order to achieve promising performance. Therefore, in the experiment on subject-independent task, we include many domain adaptation methods in the comparison. By doing so, we can effectively validate the state-of-the-art performance of our method. The comparable methods are listed as follows:
\begin{itemize}
	\item Two baseline methods: linear support vector machine (SVM)~\cite{suykens1999least}, and random forest (RF)~\cite{breiman2001random};
	
	\item Three subspace learning methods: canonical correlation analysis (CCA)~\cite{thompson2005canonical}, group sparse canonical correlation analysis (GSCCA)~\cite{zheng2016Multichannel}, and graph regularization sparse linear regression (GRSLR)~\cite{li2018eeg};
	
	\item Six transfer subspace learning methods: Kullback-Leibler importance estimation procedure (KLIEP)~\cite{sugiyama2008direct}, unconstrained least-squares importance fitting (ULSIF)~\cite{kanamori2009least}, selective transfer machine (STM)~\cite{chu2017selective}, transfer component analysis (TCA)~\cite{pan2011domain}, subspace alignment (SA)~\cite{fernando2013unsupervised}, and geodesic flow kernel (GFK)~\cite{gong2012geodesic};
	
	\item Seven recent deep learning methods: deep believe network (DBN)~\cite{zheng2015investigating}, graph convolutional neural network (GCNN)~\cite{defferrard2016convolutional}, dynamical graph convolutional neural network (DGCNN)~\cite{song2018eeg}, domain adversarial neural networks (DANN)~\cite{ganin2016domain}, bi-hemisphere domain adversarial neural network (BiDANN)~\cite{li2018novel}, EmotionMeter~\cite{zheng2018emotionmeter}, and attention-long short-term memory (A-LSTM)~\cite{song2019mped}.
\end{itemize}
All the methods are representative ones in the previous studies of emotion recognition. We directly quote (or reproduce) their results from the literature to ensure a convincing comparison with the proposed method.

The results are summarized in Table~\ref{Table: dep} and \ref{Table: ind}. Note that the subspace based methods, such as TCA, SA and GFK, are problematic to handle a large amount of EEG data due to the computer memory limitation and computational issue. Therefore, to compare with them, we have to randomly select 3000 EEG feature samples from the training data set to train these methods. Besides, the comparable methods adopting domain adaptation technique train the model with labeled training data and unlabeled test data as TANN does. From Table~\ref{Table: dep} and \ref{Table: ind}, we have three observations:
\begin{enumerate}
	\item[(1)] The proposed TANN model outperforms all the comparable methods on all the three datasets. Especially on SEED-IV dataset, the mean improvement is about 3.4$\%$ and 2.5$\%$ over the state-of-the-art methods A-LSTM and BiDANN. It verifies the learned transferable data representation are useful for EEG emotion recognition. 
	
	\item[(2)] The proposed TANN is superior to the recent domain adaptation methods. TANN has an improvement of 1.0$\%$, 3.7$\%$ and 2.1$\%$ for subject-dependent task in Table~\ref{Table: dep}, and 1.2$\%$, 2.4$\%$ and 2.5$\%$ for subject-independent task in Table~\ref{Table: ind} than the BiDANN method, which also adopts domain adversarial learning strategy to train the model. This reveals that the local and global attention structures are helpful to learn the discriminative information for emotion recognition.
	
	\item[(3)] Even under the same classification models, the performance of the subject-independent tasks are quite lower than the subject-dependent ones. It is clear to see the gaps on three datasets are about 13$\%$, 5$\%$ and 12$\%$, respectively. This reveals that the individual difference is a negative influence on EEG emotion recognition, and should be mitigated in the subject-independent task.
\end{enumerate}
\begin{table}[htb]
	\caption{The classification performance for subject-dependent EEG emotion recognition on SEED, SEED-IV and MPED datasets.}
	\centering
	\renewcommand{\arraystretch}{1.3}
	\begin{threeparttable}		
		\begin{tabular}{|c|c|c|c|} 
	\hline
	\multirow{2}{*}{\textbf{Method}} & \multicolumn{3}{c|}{\textbf{ACC / STD (\%)}} \\ \cline{2-4}
	&  SEED            & SEED-IV          &   MPED\\ \hline
	SVM~\cite{suykens1999least}          & 83.99/09.72      & ~56.61/20.05$^*$&~32.39/09.53$^*$\\ \hline
	RF~\cite{breiman2001random}           & 78.46/11.77      & ~50.97/16.22$^*$&~23.83/06.82$^*$\\ \hline
	CCA~\cite{thompson2005canonical}          & 77.63/13.21      & ~54.47/18.48$^*$&~29.08/07.96$^*$\\ \hline
	GSCCA~\cite{zheng2016Multichannel}        & 82.96/09.95      & ~69.08/16.66$^*$&~36.78/07.76$^*$\\ \hline
	DBN~\cite{zheng2015investigating}          & 86.08/08.34      & ~66.77/07.38$^*$&~35.07/11.25$^*$\\ \hline
	GRSLR~\cite{li2018eeg}        & 87.39/08.64      & ~69.32/19.57$^*$&~34.58/08.41$^*$\\ \hline
	GCNN~\cite{defferrard2016convolutional}         & 87.40/09.20      & ~68.34/15.42$^*$&~33.26/06.44$^*$\\ \hline
	DGCNN~\cite{song2018eeg}        & 90.40/08.49      & ~69.88/16.29$^*$&~32.37/06.08$^*$\\ \hline
	DANN~\cite{ganin2016domain}         & 91.36/08.30      & ~63.07/12.66$^*$&~35.04/06.52$^*$\\ \hline
	BiDANN~\cite{li2018novel}       & 92.38/07.04      & ~70.29/12.63$^*$&~37.71/06.04$^*$\\ \hline
	EmotionMeter~\cite{zheng2018emotionmeter}     & $-$              & 70.59/17.01     &$-$             \\ \hline
	A-LSTM~\cite{8606087}       &~88.61/10.16$^*$  & ~69.50/15.65$^*$&~38.99/07.53$^*$\\ \hline
	TANN    &\textbf{93.34/06.64} & \textbf{73.94/13.65} & \textbf{39.82/07.98}   \\ \hline
\end{tabular}
		\begin{tablenotes}[para]
			\footnotesize $*$ indicates the experiment results obtained are based on our own implementation.\\
			$-$ indicates the experiment results are not reported on that dataset.
		\end{tablenotes}
	\end{threeparttable}
	\label{Table: dep}
\end{table}
\begin{table}[htb]
	\caption{The classification performance for subject-independent EEG emotion recognition on SEED, SEED-IV and MPED datasets.}
	\centering
	\renewcommand{\arraystretch}{1.3}
	\begin{threeparttable}		
			\begin{tabular}{|c|c|c|c|}
	\hline
	\multirow{2}{*}{\textbf{Method}} & \multicolumn{3}{c|}{\textbf{ACC / STD (\%)}} \\ \cline{2-4}
	&    SEED    & SEED-IV        & MPED\\ \hline
	KLIEP~\cite{sugiyama2008direct}      & 45.71/17.76     &~31.46/09.20$^*$&~18.92/04.54$^*$\\ \hline
	ULSIF~\cite{kanamori2009least}       & 51.18/13.57     &~32.99/11.05$^*$&~19.63/03.81$^*$\\ \hline
	STM~\cite{chu2017selective}          & 51.23/14.82     &~39.39/12.40$^*$&~20.89/03.62$^*$\\ \hline
	SVM~\cite{suykens1999least}          &   56.73/16.29   &~37.99/12.52$^*$&~19.66/03.96$^*$\\ \hline
	TCA~\cite{pan2011domain}             &   63.64/14.88   &~56.56/13.77$^*$&~19.50/03.61$^*$\\ \hline
	SA~\cite{fernando2013unsupervised}   &   69.00/10.89   &~64.44/09.46$^*$&~20.74/04.17$^*$\\ \hline
	GFK~\cite{gong2012geodesic}          &   71.31/14.09   &~64.38/11.41$^*$&~20.27/04.34$^*$\\ \hline
	A-LSTM~\cite{8606087} 				 &~72.18/10.85$^*$ &~55.03/09.28$^*$&~24.06/04.58$^*$\\ \hline
	DANN~\cite{ganin2016domain}          &   75.08/11.18   &~47.59/10.01$^*$&~22.36/04.37$^*$\\ \hline
	DGCNN~\cite{song2018eeg}             &   79.95/09.02   &~52.82/09.23$^*$&~25.12/04.20$^*$\\ \hline
	DAN~\cite{li2018cross}               &   83.81/08.56   & 58.87/08.13    &  $-$           \\ \hline
	BiDANN~\cite{li2018novel}    & 83.28/09.60 &~65.59/10.39$^*$&~25.86/04.92$^*$  \\ \hline
	TANN                   & \textbf{84.41/08.75} & \textbf{68.00/08.35} & \textbf{28.32/05.11} \\ \hline
		\end{tabular}
		\begin{tablenotes}[para]
			\footnotesize $*$ indicates the experiment results obtained are based on our own implementation.\\
			$-$ indicates the experiment results are not reported on that dataset.
		\end{tablenotes}
	\end{threeparttable}
	\label{Table: ind}
\end{table}
\subsection{Discussion}
\subsubsection{The confusion of different emotions based on TANN model}
\begin{figure}[htb]
	\centering
	\subfigure[SEED]{
		\label{CM:SEED-dep}
		\includegraphics[width=0.375\linewidth]{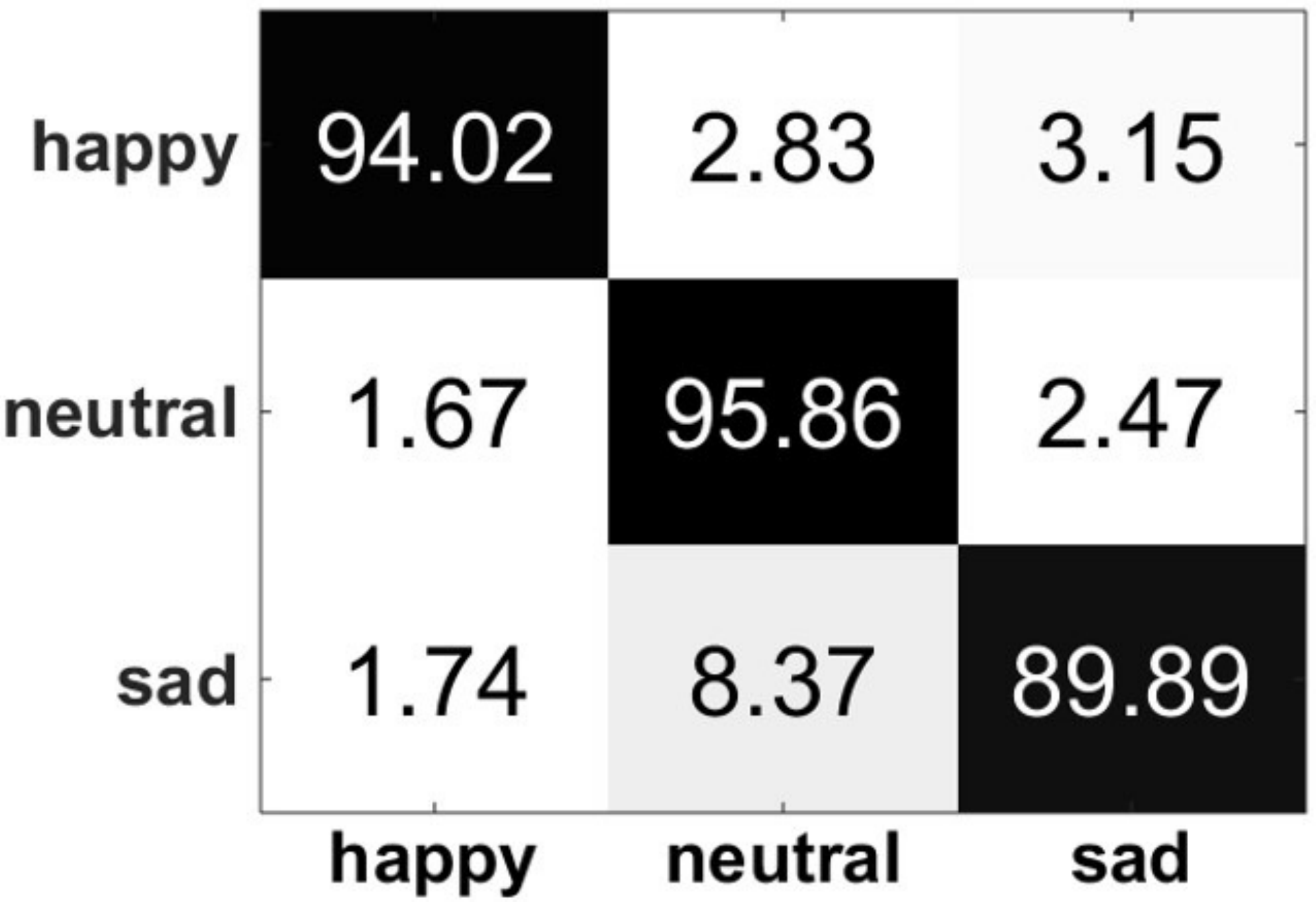}}
	\subfigure[SEED-IV]{
		\label{CM:SEED-IV-dep}
		\includegraphics[width=0.375\linewidth]{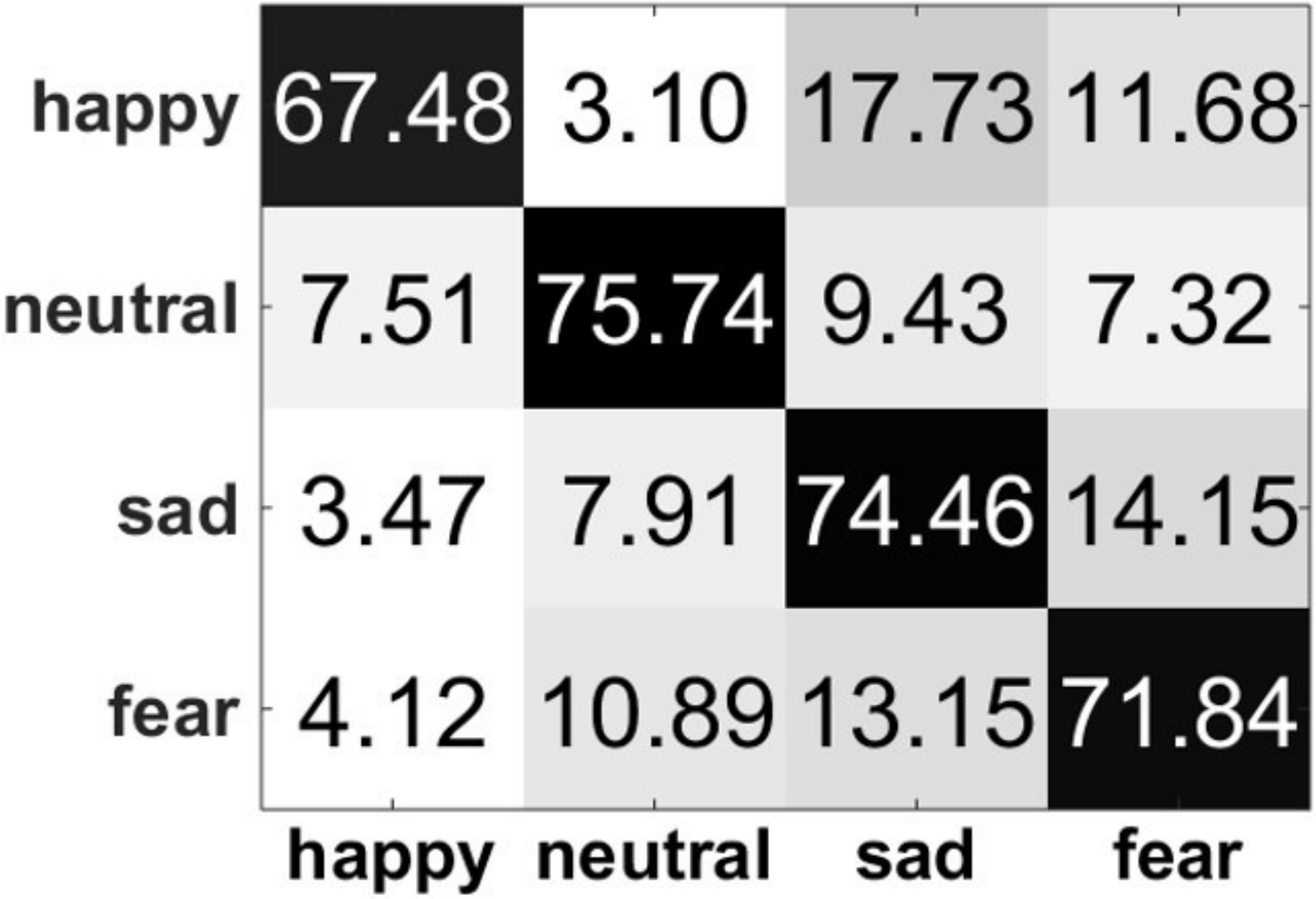}}	
	\subfigure[MPED]{
		\label{CM:MPED-dep}
		\includegraphics[width=0.81\linewidth]{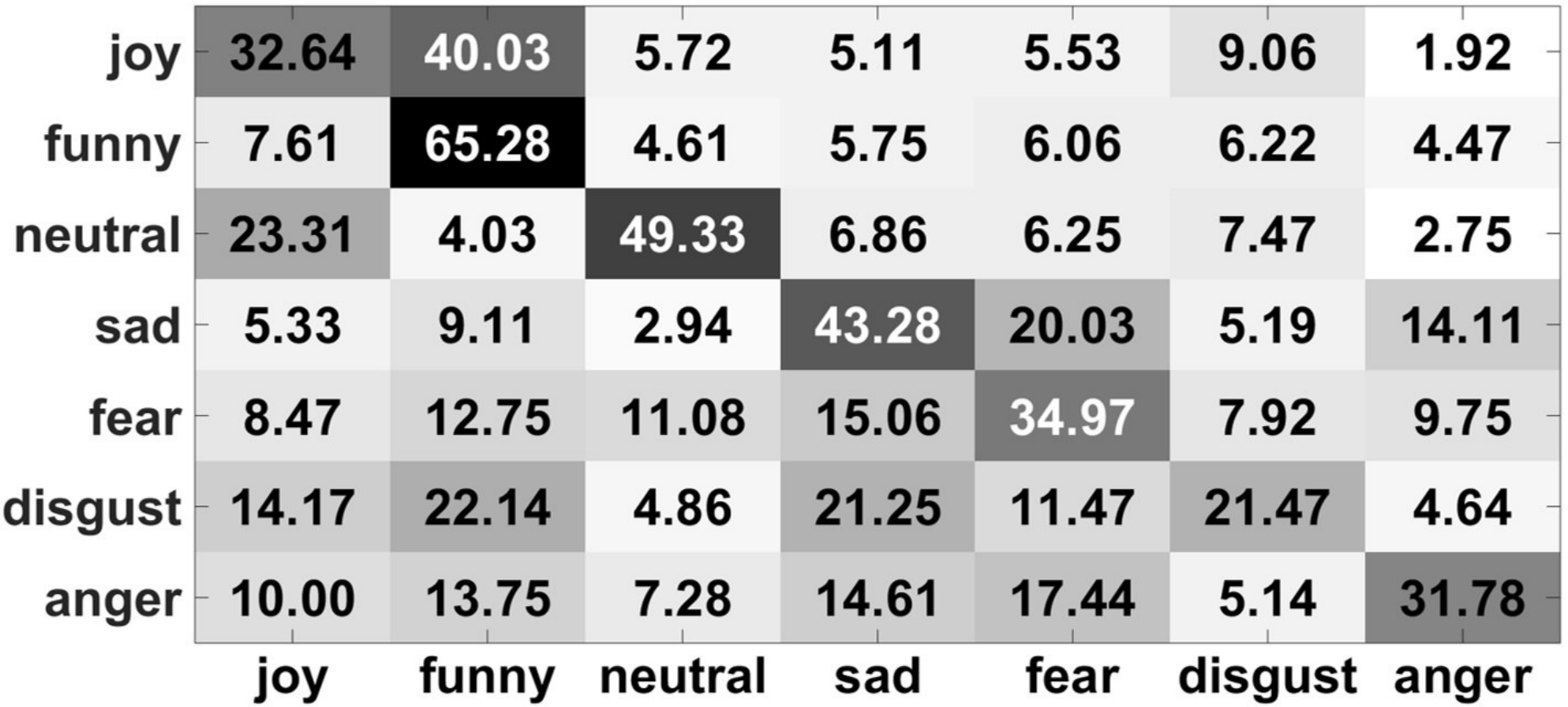}}
	\caption{The confusion matrices based on the subject-dependent experimental results on three datasets.}
	\label{Fig: CM1}
\end{figure}

\begin{figure}[htb]
	\centering
	\subfigure[SEED]{
		\label{CM:SEED-ind}
		\includegraphics[width=0.375\linewidth]{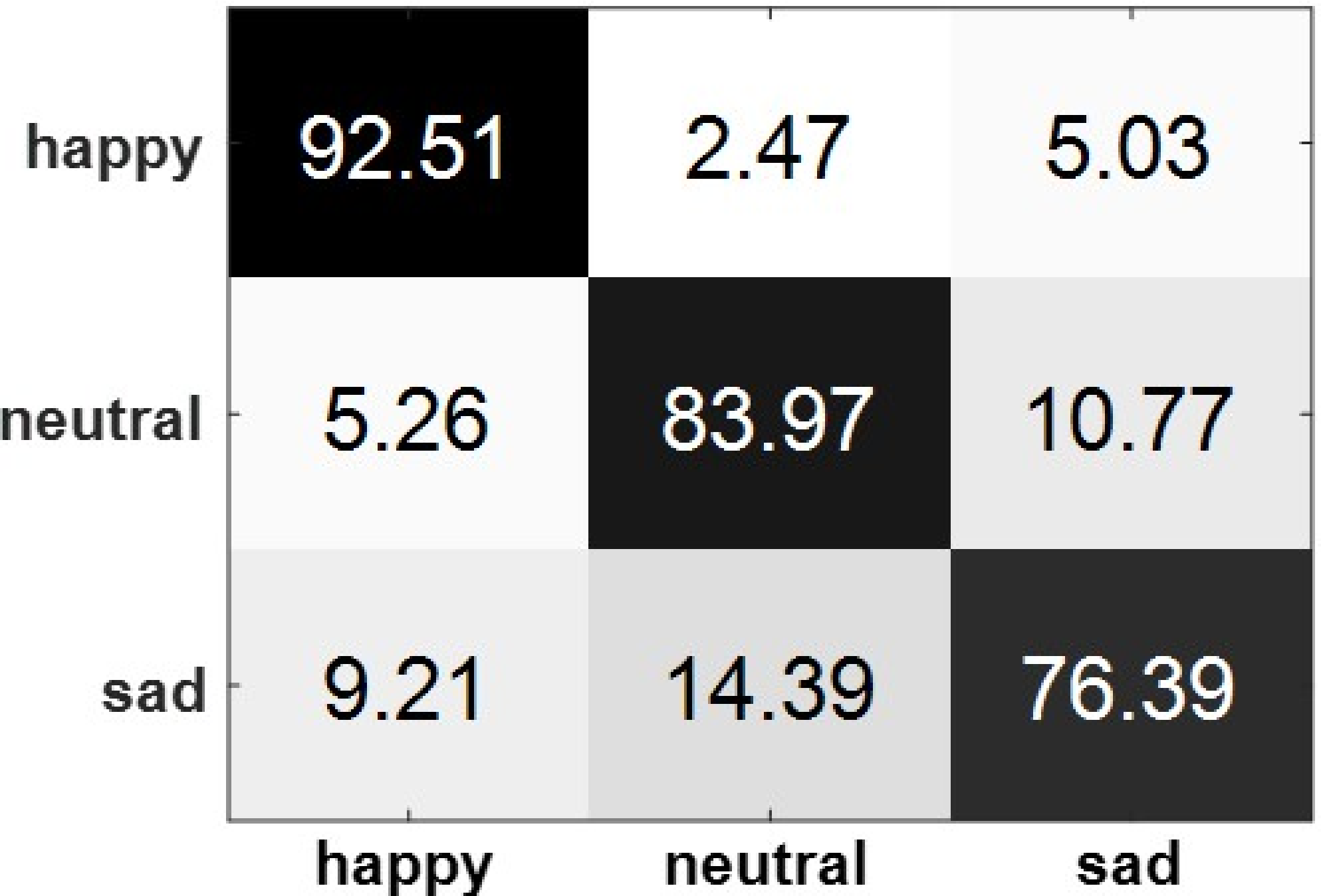}}
	\subfigure[SEED-IV]{
		\label{CM:SEED-IV-ind}	
		\includegraphics[width=0.375\linewidth]{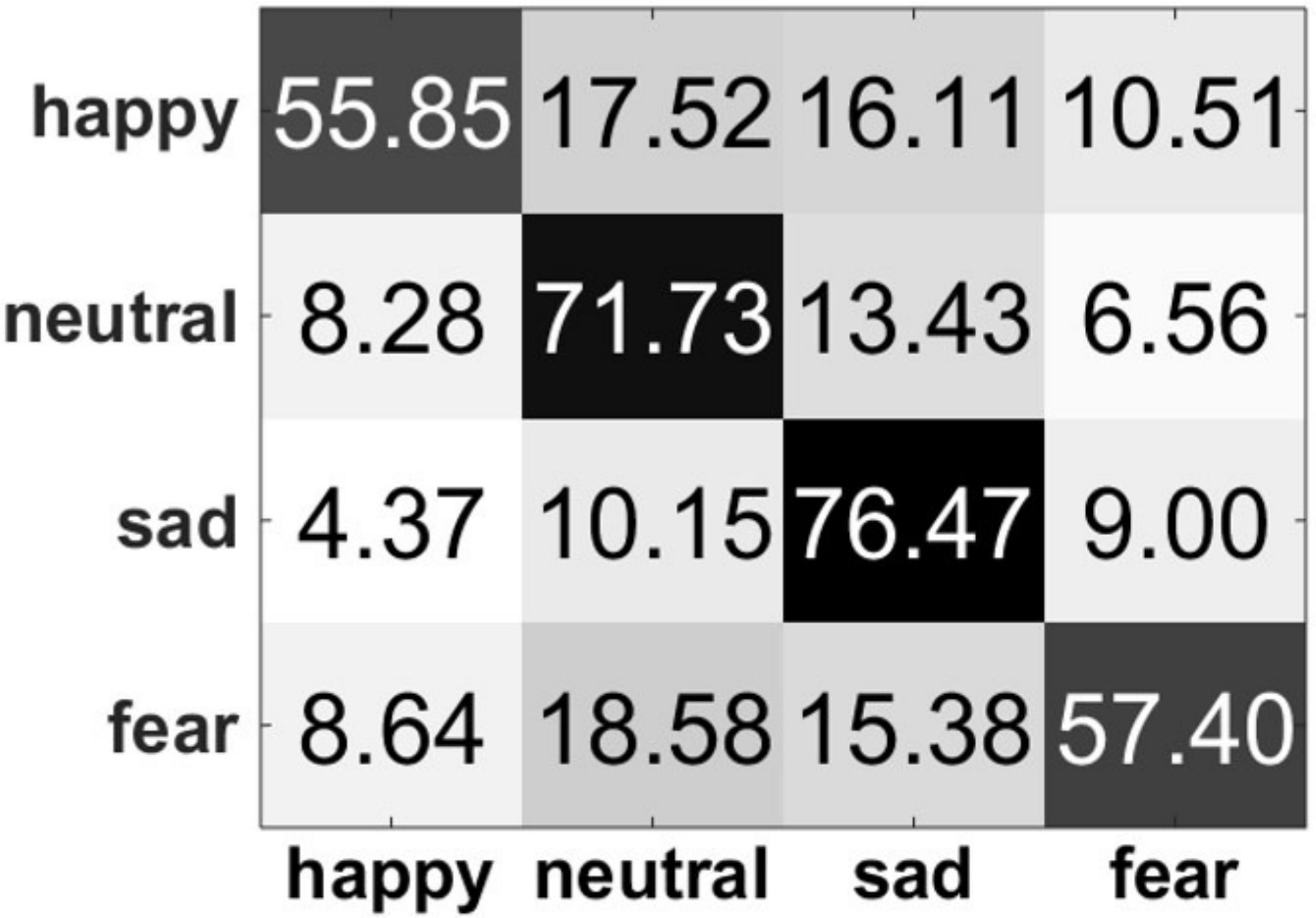}}
	\subfigure[MPED]{
		\label{CM:MPED-ind}
		\includegraphics[width=0.81\linewidth]{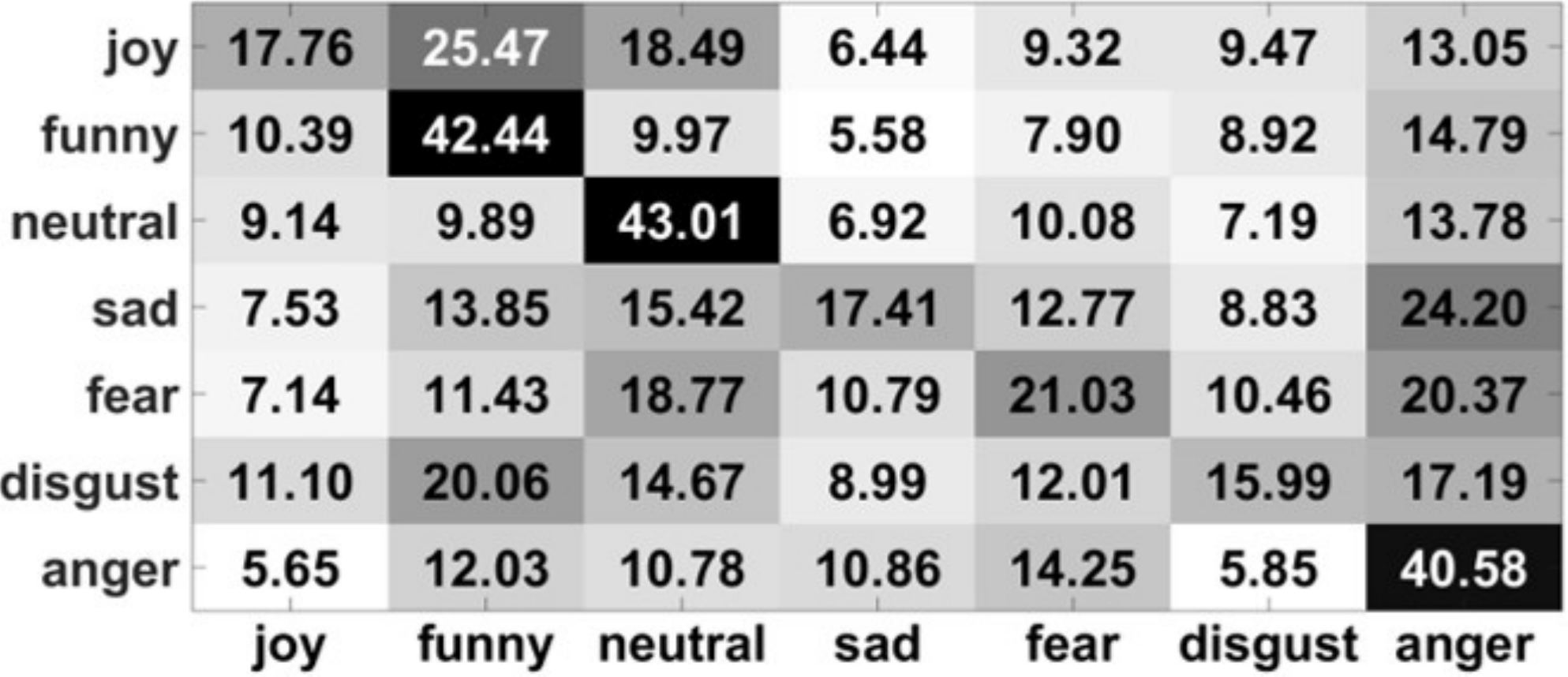}}	
	\caption{The confusion matrices based on the subject-independent experimental results on three datasets.}
	\label{Fig: CM2}
\end{figure}

To better understand the confusion of TANN in recognizing different emotions, we depict the confusion matrices of subject-dependent and subject-independent EEG emotion recognition experiments in Fig.~\ref{Fig: CM1} and~\ref{Fig: CM2}, respectively, from which we have the following observations:
\begin{enumerate}
	\item[(1)] In Fig.~\ref{Fig: CM1}, for SEED, the classification accuracies for three emotions are about 90$\%$, and the happy and neutral emotions are easier to be recognized than the sad emotion; for SEED-IV, which consists of four emotions, we can see the negative emotions, i.e., sad and fear, are confused by the classifier with higher possibility; and for MPED, the confusion is more complex because it has more emotions than the other two datasets. It is obvious to see that the funny emotion is the easiest to be recognized and has 16$\%$ more than the neutral emotion on the second place. Except this, we can find that the funny and joy are easier to be confused maybe because both of them are positive emotions.	
	\item[(2)] From the results of subject-independent EEG emotion recognition experiment in Fig.~\ref{Fig: CM2}, we can observe that, for SEED, which has three types of emotions, the happy emotion is much easier to be recognized than neutral and sad; for SEED-IV, the neutral and sad emotions are much easier to be recognized; for MPED, which is a hard seven classification problem, the accuraries of funny, neutral and anger emotions overpass that of the other emotions, and this reveals that we should focus on the joy, sad, fear and disgust emotion data in the task of classifying seven emotions.	
\end{enumerate}

\subsubsection{The transferability of different brain regions}
\begin{figure}[htb]
	\centering
	\subfigure[SEED]{
		\includegraphics[width=0.295\columnwidth]{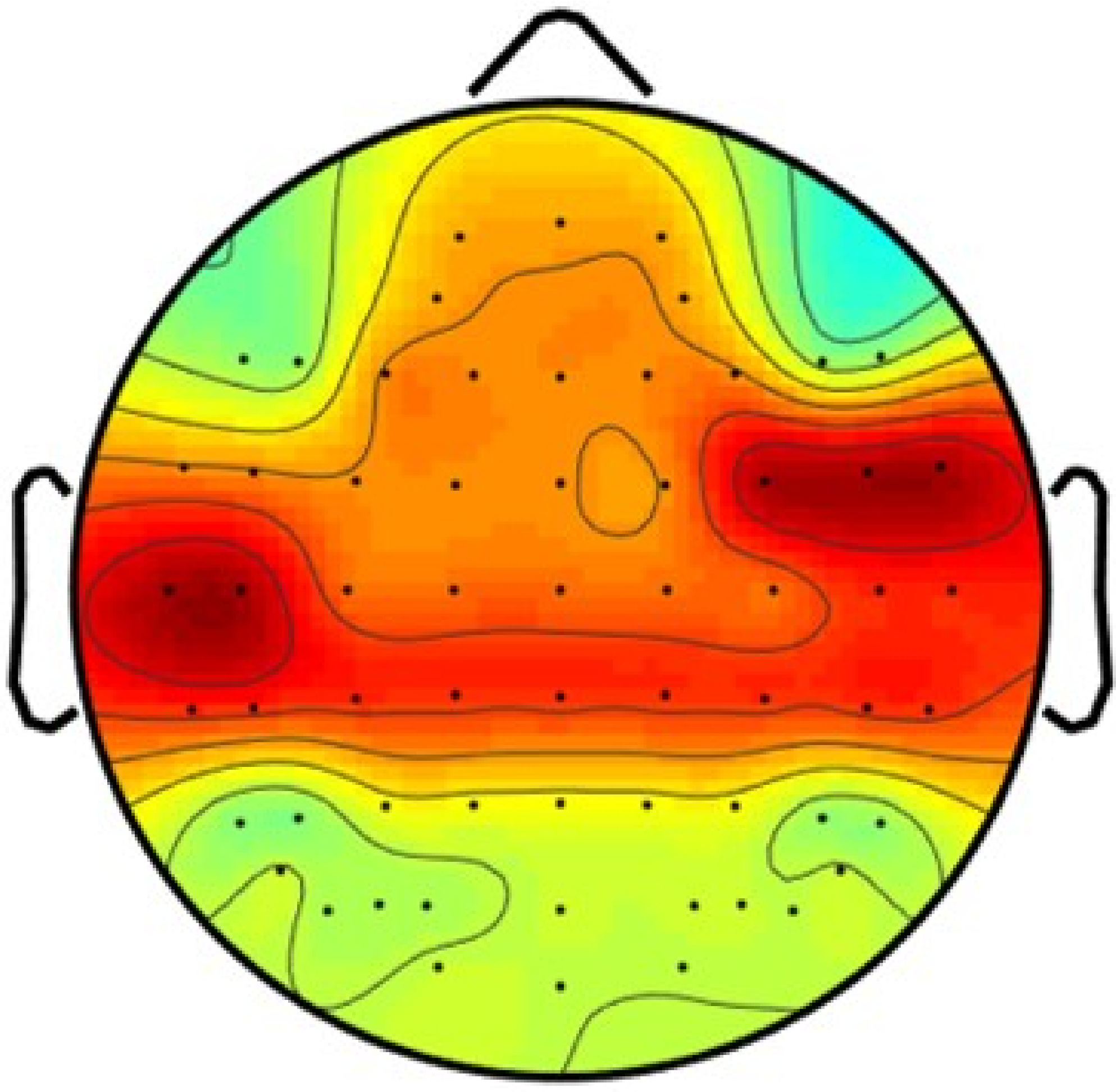} }
	\subfigure[SEED-IV]{
		\includegraphics[width=0.295\columnwidth]{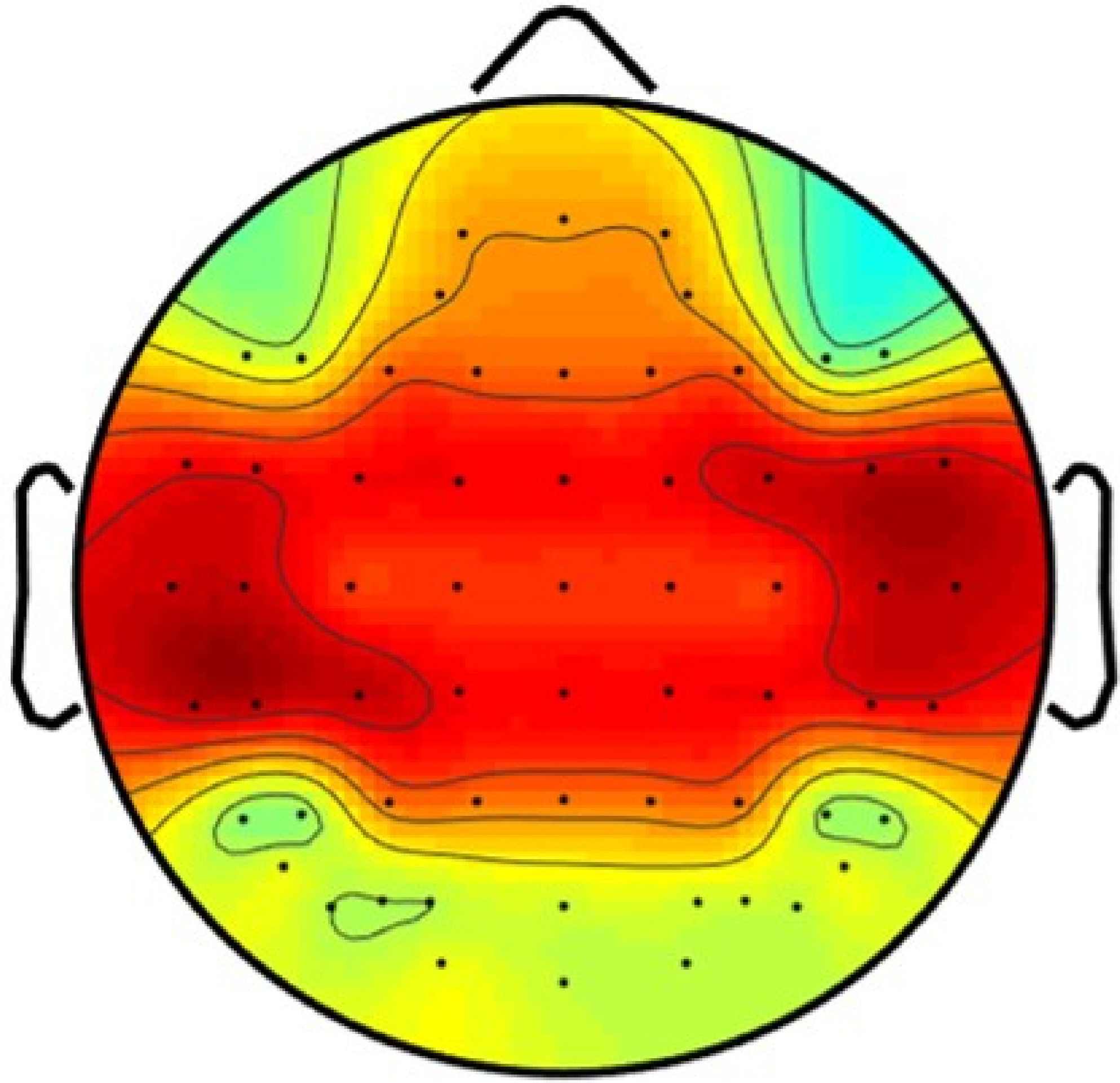} }
	\subfigure[MPED]{
		\includegraphics[width=0.295\columnwidth]{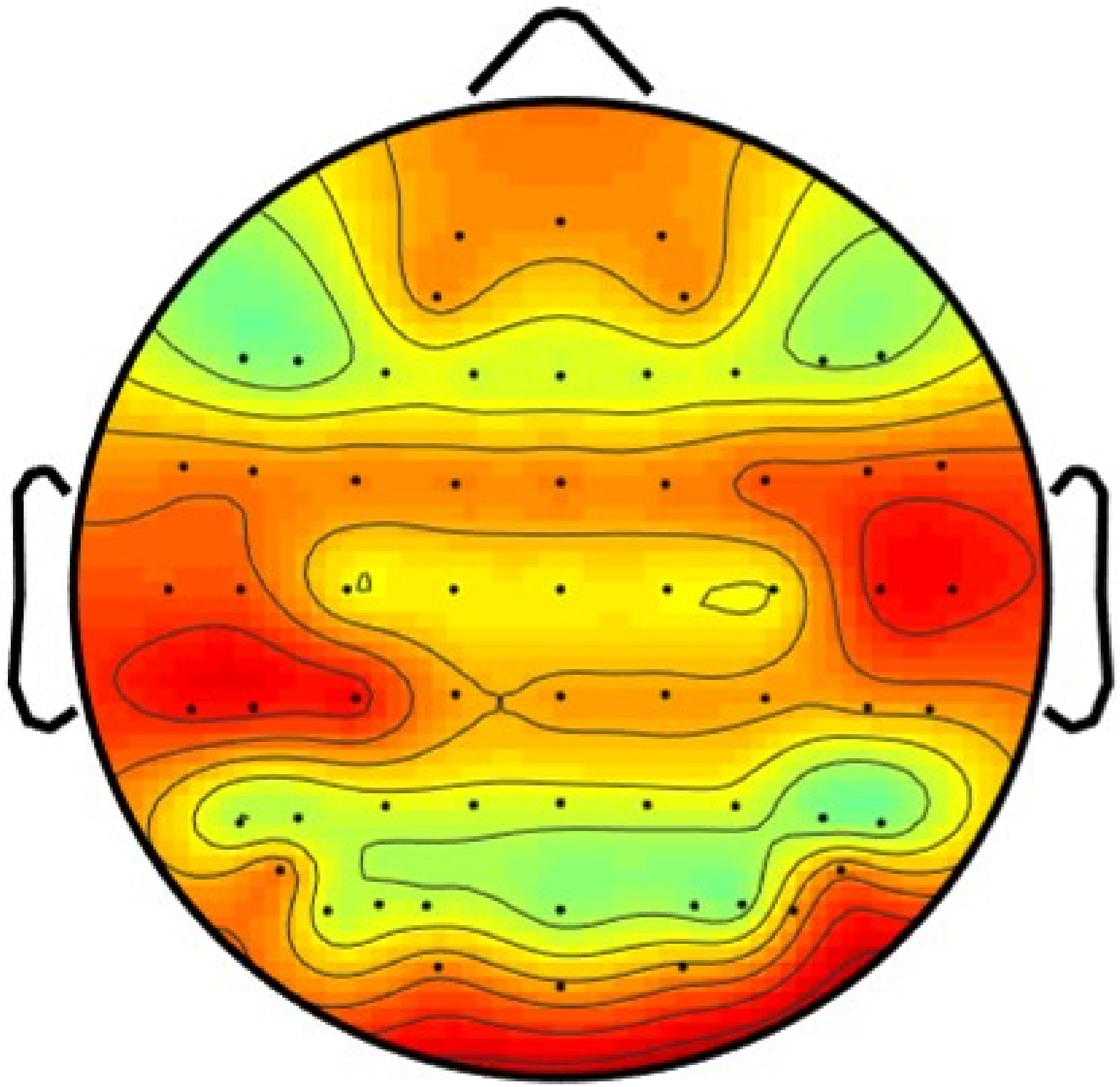} }
	\caption{The transferability of different EEG brain regions.}
	\label{Fig: Transferability_map}
\end{figure}

To investigate the transferability of different brain regions for EEG emotion recognition, we visualize all the brain regions by mapping the local attention values $w$ in Eq.~(\ref{Eq: 1-Entropy}) into the corresponding electrodes. The obtained results are shown in Fig.~\ref{Fig: Transferability_map}, from which we have two observations:
\begin{enumerate}
	\item[(1)] The left and right temporal lobes make more important contribution for emotion recognition in all the three datasets, which coincides with the previous EEG emotion studies~\cite{lin2010eeg,zheng2015investigating}. This also reveals that, as well as the proposed model can adaptively give attention to different brain regions, it is still effective to capture the most important ones.
	\item[(2)] The activation areas are slightly different across datasets. For example, there is a broader activation to the temporal lobes for SEED-IV compared with SEED. And for MPED, which consists of more types of emotions, the occipital lobe, as well as the temporal lobe, contributes more for emotion expression.
\end{enumerate}

\subsubsection{Ablation study}
\begin{figure}[htb]
	\centering
	\subfigure[TANN-R1]{
		\includegraphics[width=0.95\columnwidth]{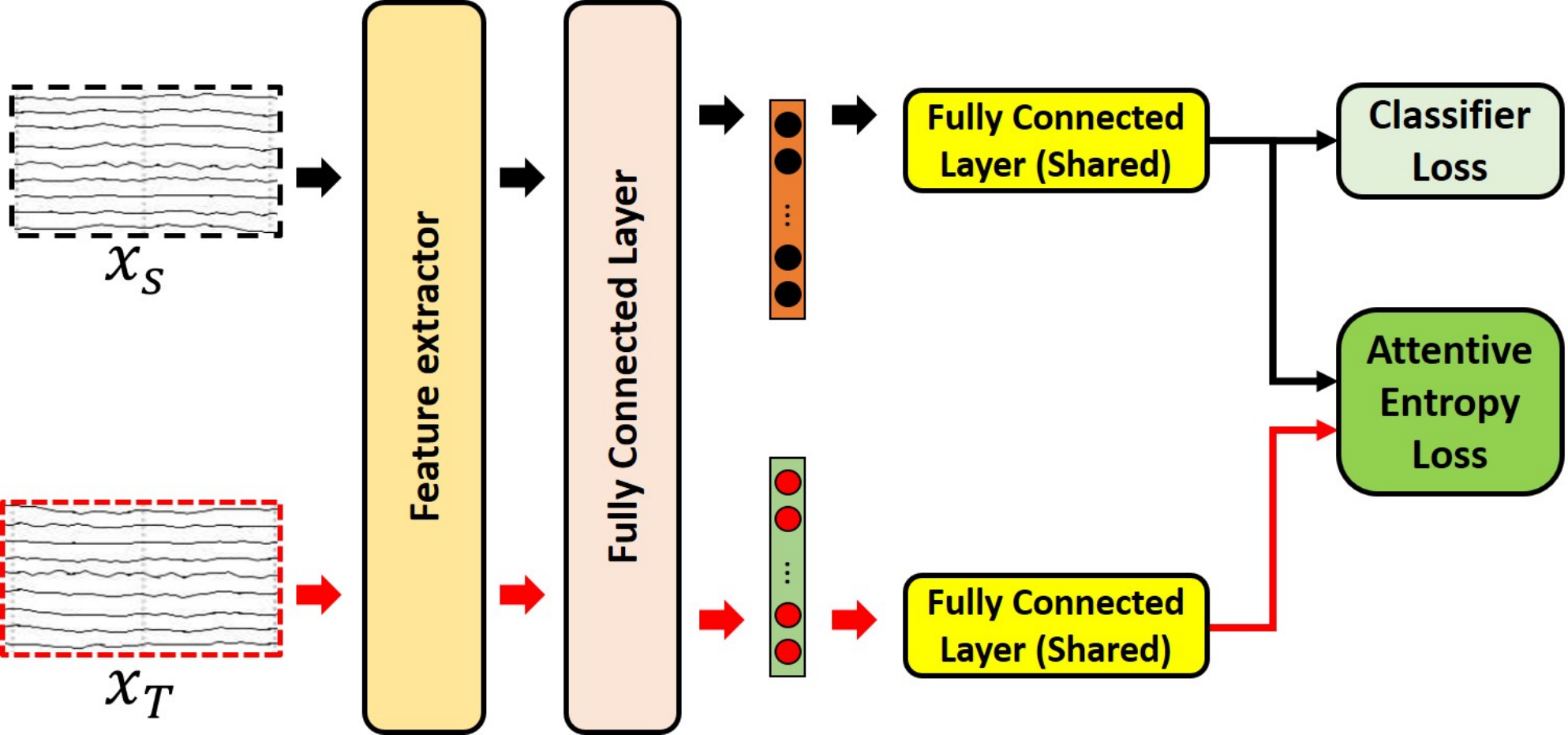} }
	\subfigure[TANN-R2]{
		\includegraphics[width=0.95\columnwidth]{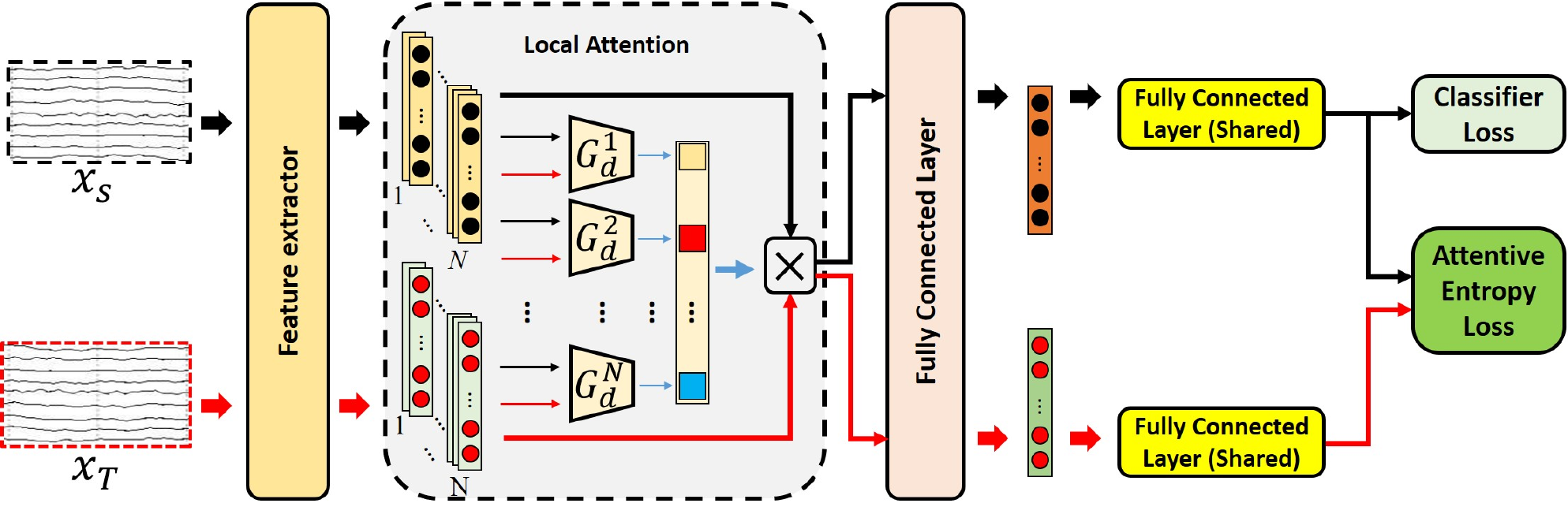} }
	\subfigure[TANN-R3]{
		\includegraphics[width=0.95\columnwidth]{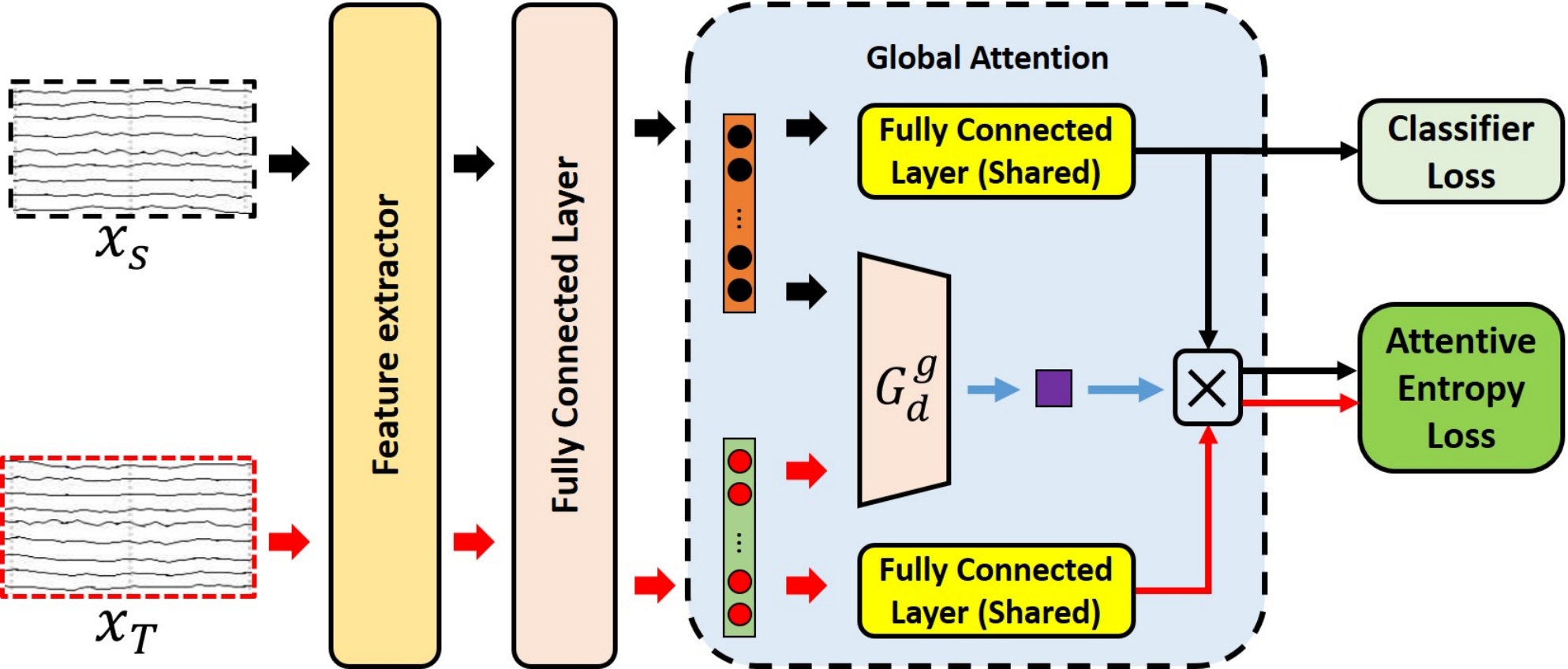} }
	\caption{The frameworks of the reduced models of TANN: (a) TANN-R1, (b) TANN-R2, (c) TANN-R3.}
	\label{Fig: Ablation study}
\end{figure}

To see the importance of each module of TANN for EEG emotion recognition, we conduct an ablation study by removing the local and global attention layers both and separately. These reduced models are depicted in Fig.~\ref{Fig: Ablation study}, which includes
\begin{itemize}
	\item TANN-R1, which removes both the local and global attention modules;
	\item TANN-R2, which neglects the global transferability for EEG samples;
	\item TANN-R3, which employs the same structure of TANN model except the local attention layer.
\end{itemize}

The experimental results are shown in Table~\ref{Table:The ablation study}, from which we can have three observations:
\begin{enumerate}
	\item[(1)] It is effective for the structure of the feature extractor in TANN. From the results of TANN-R1, we can see it achieves comparable performance on three datasets. This verifies the obtained deep data representation by two directional recurrent neural networks is discriminative for emotion recognition.
	
	\item[(2)] Either the local or global transferable attention modules can enhance emotion recognition. In contrast to TANN-R1, TANN-R2 and TANN-R3 improve the accuracy, on average, by 1.8$\%$ and 1.5$\%$ on three datasets, respectively. 
	
	\item[(3)] By assembling the feature extractor, local and global attention modules, TANN achieves the best performance. We can see TANN has a further improvement of 3$\%$ compared with TANN-R2 and TANN-R3.
\end{enumerate}
The above results verify the effectiveness of the three important modules in TANN.

\begin{table}[htb]
	\caption{The comparison of EEG emotion recognition results among four methods: (1) TANN-R1, (2) TANN-R2, (3) TANN-R3; (4) TANN.}
	\centering
	\renewcommand{\arraystretch}{1.3}
		\begin{threeparttable}		
			\begin{tabular}{|c|c|c|c|} 
				\hline
				\multirow{2}{*}{\textbf{Method}} & \multicolumn{3}{c|}{\textbf{ACC / STD (\%)}}\\ \cline{2-4} 
				&  SEED            & SEED-IV          &   MPED  \\\hline
				TANN-R1    &  87.06/09.45       &   68.28/14.28   &  37.92/07.80   \\\hline
				TANN-R2    &  89.73/07.53       &   \textbf{70.82/14.65}   &  \textbf{38.10/07.98}   \\\hline
				TANN-R3    &  \textbf{91.03/07.63}       & 68.72/13.30     &  38.06/08.21   \\\hline
				TANN    &\textbf{93.34/06.64} & \textbf{73.94/13.65} & \textbf{39.82/07.98} \\\hline
			\end{tabular}
			\begin{tablenotes}[para]
			\end{tablenotes}
		\end{threeparttable}
	\label{Table:The ablation study}
\end{table}

\section{Conclusion}
\label{Sec: Conclusion}
In this paper, we propose a transferable attention neural network (TANN) to deal with EEG emotion recognition problem, which is motivated by the finding that not all the training samples have the equal contribution for emotion recognition, which also happens for the importance of different brain regions in this sample. TANN has the ability to learn the positive and negative information from the sample-level and brain-region-level, which can improve EEG emotion recognition. The proposed framework is easy to implement and the extensive experiments on three public EEG emotion datasets demonstrated that the proposed TANN method achieves the state-of-the-art performance. Besides, based on TANN, we also investigate the transferability of different brain regions in EEG emotion recognition and find that the temporal lobe and occipital lobe contribute more for emotion expression. In the future work, we will further investigate more operations for learning the transferability information to explore the potential efficacy of transferable attention for EEG emotion recognition.

\ifCLASSOPTIONcaptionsoff
\newpage
\fi

\bibliographystyle{IEEEtran}
\bibliography{manuscript_bib}

\end{document}